\documentclass{article}

\usepackage{graphicx}
\usepackage{subfigure}

\PassOptionsToPackage{sort&compress,numbers}{natbib}
\usepackage[final]{neurips_2024}
\usepackage{framed}
\usepackage[utf8]{inputenc} 
\usepackage[T1]{fontenc}    
\usepackage{hyperref}       
\usepackage{url}            
\usepackage{booktabs}       
\usepackage{amsfonts}       
\usepackage{nicefrac}       
\usepackage{microtype}      
\usepackage{xcolor}         

\title{Protocol Learning, Decentralized Frontier Risk and the No-Off Problem}

\author{%
  Alexander Long \\
  Pluralis Research\\
  \texttt{alexander@pluralis.ai}} 

\begin{document}
\maketitle

\begin{abstract}
Frontier models are currently developed and distributed primarily through two channels: centralized proprietary APIs or open-sourcing of pre-trained weights. We identify a third paradigm—\emph{Protocol Learning}—where models are trained across decentralized networks of incentivized participants. This approach has the potential to aggregate orders of magnitude more computational resources than any single centralized entity, enabling unprecedented model scales and capabilities. However, it also introduces novel challenges: heterogeneous and unreliable nodes, malicious participants, the need for unextractable models to preserve incentives, and complex governance dynamics. To date, no systematic analysis has been conducted to assess the feasibility of Protocol Learning or the associated risks, particularly the “No-Off Problem” arising from the inability to unilaterally halt a collectively trained model. We survey recent technical advances that suggest decentralized training may be feasible—covering emerging communication-efficient strategies and fault-tolerant methods—while highlighting critical open problems that remain. Contrary to the notion that decentralization inherently amplifies frontier risks, we argue that Protocol Learning’s transparency, distributed governance, and democratized access ultimately reduce these risks compared to today’s centralized regimes.

\end{abstract}

\section{Introduction} \label{sec:intro}
Fundamental modeling advances \citep{transformer} combined with unprecedented data and compute scale have resulted in models able to perform routine knowledge work to a level beyond the standard human \citep{sparks, gpt3, gpt4}. Current trends give every indication that increasing scale will continue to increase model performance \citep{scalinglaws,predictablescaling}. Calls for pauses to this line of research have, to date, been entirely unsuccessful \citep{pauseletter}. While continued increase in model capability seems likely with further scale, it is not clear at which point such models gain the ability to solve certain tasks \citep{emergentprops,emergentmirage}, or if they will at all. Consequently despite widespread understanding of the misalignment and misuse risks such models pose, because the front of tasks being solved has clear economic utility, there is an ongoing race to train larger models at larger and larger scale. Due to such high cost, we appear to be trending towards an oligopoly of frontier model providers. In this scenario, the capital expenditure required for effective training and deployment results in a barrier that prevents all but a handful of large corporations and nation-states from training such models. 

A prevailing view is opensource models will provide a counterbalance to the emerging frontier model oligopoly. This ignores the huge cost of training such models and the inability of opensource to monetize the models and hence recoup this cost. Releasing the output of a process that requires hundreds of millions to billions of dollars of cost for free, without constraints, is unsustainable and will not continue. Releasing the weights of early models is a good strategy to drive adoption and developer mind-share when lagging in overall model capability, however at some point monetization must begin to provide a return. There may be many forms of this aside from simply charging for model access, however the terminal state is that some value will be extracted from model consumers which, in aggregate, will be greater than the total cost to train. 

Protocol Learning is as a third alternative to the current centralized and open-source approaches. In this paradigm, models are trained collaboratively across a network of incentivized participants, each contributing compute resources in exchange for partial ownership. This approach addresses two critical limitations of current approaches: it removes the dependency on centralized entities releasing their trained models to the public, and it enables smaller participants to collectively achieve the massive scale required for frontier model training through resource pooling. However, assembling sufficient compute power for genuine frontier model training requires explicit incentives – voluntary contribution alone cannot match centralized training capabilities. While these incentives are necessary to achieve competitive scale, they fundamentally alter the risk landscape of AI development by introducing novel governance challenges and safety concerns.

Our analysis proceeds in four parts. First, we examine computational capacity across three training paradigms: non-incentivized collaborative networks, centralized clusters, and incentivized protocols. We demonstrate that properly incentivized protocols can mobilize computing resources that exceed centralized approaches by orders of magnitude (Sec. \ref{sec:compute}). Second, we survey recent advances in decentralized training and inference, revealing that the technical foundations for Protocol Learning are more mature than commonly recognized (Sec. \ref{sec:feasublity}). Third, we propose a framework for incentivization through fractional model ownership, analyzing both the system properties this implies and critical open research challenges (Sec. \ref{sec:main}). Finally, we present the first comprehensive analysis of Decentralized Frontier Risk (DFR), examining how Protocol Learning fundamentally reshapes the risk landscape of advanced AI development (Sec. \ref{sec:frontierrisk}). In addition to identifying several novel risks — most notably the "No-Off Problem" — our analysis suggests that Protocol Learning's increased transparency and democratized access reduces overall frontier risk.

\section{Comparing Centralized, Decentralized, and Incentivized Compute Capacity}\label{sec:compute}
The scale of centrally controlled compute capacity is perhaps underappreciated; it is enormous today and growing rapidly. Precise figures are not publicly disclosed; however as a single public datapoint, Meta has announced plans to purchase 350k H100s by year end 2024 \citep{fbpurchase} - on the order of 350 exaFLOPS at theoretical peak load using TF32 datatype \citep{h100specs}. 

In contrast, the maximum compute capacity achieved by volunteer networks was a temporary peak of 1.2 full-precision exaFLOPS by the Folding at Home Project \citep{foldingathome} in March 2020. Over 2 million devices (1.4M of which were CPU's) were present in the swarm, triggered by a surge of interest in projects simulating theoretically druggable protein targets from SARS-CoV-2. In short, volunteer network capacity peaked at two orders of magnitude below a single centralized actors compute purchases in a single year.

Compared to volunteer networks and centralized clusters, incentivized decentralized swarms, such as those assembled for Proof-of-Work (PoW) mining in the Bitcoin \citep{bitcoin} and Ethereum \citep{ethereum} protocols have achieved orders of magnitude larger capacity than any centralized cluster. We measure productive capacity here in terms of Watts rather than FLOPS in order to make meaningful comparisons (PoW mining does not involve any floating point operations). 350k H100's running continually at peak power draws 0.24 GW (1 GW is the average energy consumption rate of a one-million inhabitant industrial city). In contrast; Bitcoin PoW mining consumption is estimated at $150\pm 50$ TWh in 2022 \citep{bitcoinenergy}, or 17.12 GW on average and approximately 0.5\% of total worldwide energy consumption. While these figures are approximate, the core fact remains striking; given certain incentives, pooled compute two orders of magnitude larger than the already enormous largest pools of centrally controlled compute can, and has, been assembled under a single protocol. 

We also note that while decentralized compute hardware cannot be used to contribute the centralized models the reverse is not true. High capacity centralized clusters can be used to participate in training and hence ownership of decentralized models. 

\section{Feasibility of Decentralized Training} \label{sec:feasublity}
A common objection to large-scale decentralized training over the internet is that it is infeasible given the node-node communication speeds required for frontier model training are orders of magnitude slower than those in centralized clusters (i.e. 100 MB/s standard internet vs. 200GB/s infiniband). However, recent work (Sec. \ref{overhead}) has proposed approaches to robust decentralized training that are extremely communication efficient. 

Simply being communication efficient is not enough; we define non-incentivized, large-scale decentralized training as distributed training (i.e. multi-node training) with the following additional properties; 
\begin{enumerate}
    \item Communication efficient: does not require node-to-node bandwidth and latency above that of standard internet connections. 
    \item Model sharding: models much larger than can fit on any single node are able to be trained. 
    \item Elastic training: any node can drop or enter the swarm at will without disrupting training. 
    \item Byzantine tolerance: even with a fraction of nodes being actively malicious, training cannot be disrupted. This in turn implies no master node or parameter server. 
    \item Heterogeneous node capacity: nodes can have varying compute power, communication bandwidth and latency.
\end{enumerate}

While current solutions demonstrate many of these properties individually with minimal overhead compared to centralized training, no existing system combines all five. The challenge lies not just in implementing each property, but in their interaction—Byzantine fault tolerance becomes more complex with elastic participation, heterogeneous compute complicates model sharding, and so on. Finally, even if achieved, such a system would enable only voluntary collaborative training. Protocol Learning requires additional mechanisms for incentivization and governance, building upon this technical foundation. Below, we examine current progress toward each property and identify key remaining challenges.

\subsection{The Problem of Communication Efficiency} 
Due to the computational intensity of Deep Learning, distributed training has been essential since it's inception \cite{dp,ps}. The simplest approach is Distributed Data Parallel (DDP) \cite{ddppytorch}, where each node contains complete copies of the model carrying out forward and backward passes on batches of data independently. At each iteration, gradients are averaged across nodes. Most DDP implementations make use of the all-reduce primitive; an operation which gathers equally sized tensors (gradients in this case), applies a reduction operation (sum, max etc.), and returns the result to each node. Several variants of All-Reduce have been proposed to minimize communication overhead \cite{hovrod,butterflyallreduce,allreducevariants}. These implementations are exact, synchronous, and have similar convergence guarantees as standard SGD, but assume homogeneous, static compute for each node, a fixed node topology, and are largely not fault tolerant. While synchronous, all-reduce DDP can be implemented without a parameter server \cite{ps} or orchestration/master node as in standard Federated Learning (FL) (see Sec. \ref{sec:FL}). To further reduce communication overhead, fixed \cite{qsgd} or adaptive \cite{adaptivequantization} compression is often applied to the gradients, reducing the amount of information that must be transferred. While this results in weaker convergence guarantees, in practice gradient compression significantly reduces walltime with only minor effect on performance \cite{qsgdsummary}. 

\textbf{DDP approaches require each node to be capable of storing and operating on the full model.} In the Protocol Learning case, where individuals are not required to have access to large clusters, this is infeasible and some form of model parallelism must be implemented to shard the model across nodes. There are several existing strategies for model parallel that are typically used in the centralized case, however such approaches are very communication intensive. The simplest model parallel implementation is pipeline parallelism \cite{pipeparrellel,megatronmodelparallel,gpipe} where the model is distributed layer-by-layer and communication is node-to-node. More common for very large models is Fully Sharded Data Parallel (FSDP) \cite{fsdp}, or ZeRO-3 \cite{zero3}, two implementations of the same concept. FSDP decomposes the model both layer-wise and splits the layers themselves across devices. Each accelerator contains only a `shard' of each layer in memory, gathering the shards required to perform the local forward and backward passes from other accelerators at the time they are required. Hence, FSDP significantly increases the size of models that can be trained, but is very communication intensive as each layer requires all-to-all communication between nodes holding that layer. Although there is early work on communication efficient FSDP \cite{qfsdp}, the communication overhead required by FSDP and related approaches is the central reason such fast node-to-node interconnects (currently approaching 1 TB/s) are required in standard large-model training. 

\subsection{Communication Efficient Training}\label{overhead}
Several works have made progress towards large-scale model training in the decentralized setting. Communication efficient DDP has been achieved with gossip protocols in place of the synchronous all reduce and a specific topology \cite{gossipproto,gossipdl}. Convergence guarantees can still be obtained in this setting \cite{D-PSGD, AD-PSGD} even with the communication graph altering during training \cite{dladaptive,unifieddsgd}. 

Decentralized training with heterogeneous devices has also been demonstrated, but constrained to small models. Moshpit-SGD \cite{moshpitsgd} introduces an approach that is both communication efficient and scales well with heterogeneous compute and communication bandwidth. \citet{opencollabs} perform a real run, over 200 MB/s interconnects, using devices with a range of capabilities on a dynamic swarm to train a 72.5M parameter ALBERT-xlarge variant. Here a community of volunteers completed a training run that would have taken over 3 months using a standard 8xV100 setup, within 22 days on consumer hardware. Nodes were verified with an allow-listing approach, all gradient computations are assumed to be correct and each node computes full model gradients and hence has a copy of model weights. 

Large models are also possible, with \citet{swarmparellel} showing pipeline parallel training becomes \textit{less} communication intensive relative to compute as models grow larger; hence improving utilization. This work practically demonstrated training of a 1B parameter LLM on T4 GPU's with 500 MB/s interconnects, achieving roughly 20\% throughput overhead to centralized training, maintaining very high accelerator utilization, and also possessing basic fault tolerance. This is achieved with redundancy within each pipeline stage, and dynamic and elastic routing between each stage, and the assumption of good actors. Learning@Home \cite{decenmoe} propose the Decentralized MoE architecture and an asynchronous training scheme in order to achieve communication efficiency over heterogeneous nodes. Such an approach can theoretically scale to very large parameter sizes but has not been practically demonstrated beyond 257M, and while node failures are handled, byzantine nodes are not.  

\subsection{Byzantine-Tolerant Training}
Decentralized Training must be robust to malicious actors, especially when incentives are present. Such bad-actors may try to sabotage training runs or alter model behaviour by providing incorrect gradients, attempt to gain control of the swarm or simply steal ownership credentials. Byzantine robustness has been studied in the centralized setting, where there are no bad actors, due to the need to reduce the impact of non-malicious failures in distributed training (e.g. network and hardware failures) which become common at large scales, and can derail an entire training run. \citet{byztraining} note that as the byzantine gradients are potentially unbounded, no linear aggregation can be byzantine robust, and propose an approach the precludes outliers, assuming them to be byzantine. This results in slower convergence but is provably byzantine robust. \citet{byzrob2,Aggregathor} improve on the convergence bounds with alternative aggregation procedures, however the effectiveness is in question as while they provably reach flat regions in the non-convex case, in practice attacks can be devised to make such models effectively useless \cite{byzrobproblems,byzrobproblems2, byzrobproblems3}. In addition, all such approaches are data parallel, and contain a single point of failure in the parameter server itself.

There is comparatively little work on byzantine robustness in the the decentralized setting. \citet{peng2021} and \citet{el2020genuinely} propose approaches and associated convergence bounds that meet Properties 1 and 3. \citet{secatscale} is the only work to prove robust convergence bounds in the non-convex loss case, without assuming a bound on the gradient variance using a combination of gossip SGD, Centered Clip aggregation \cite{centeredclip}, validate nodes and implements an accuse protocol to ensure correctness of the validators. \citet{byztraining} achieves a convergence rate equivalent to D-SGD without faults, however assumes each node has access to the entire training dataset; this allows byzantine nodes to be detected by checking computation explicitly. \citet{monna} considers an even more general setting where it is not necessary for each node to have access to the entire dataset and proves a convergence rate that is linear with the number of byzantine nodes, up to 9.09\%. Such approaches introduce surprisingly little overhead over the non-byzantine case, but do not generalize to sharded training.

\begin{framed}
In summary, many of the foundational elements of Protocol Learning are more mature than generally recognized by the research community. Key capabilities have been demonstrated independently: communication-efficient distributed data parallel training \citep{gossipproto,gossipdl}, billion-parameter model training using pipeline parallelism over standard internet connections \citep{swarmparellel}, and theoretical frameworks for byzantine fault tolerance \citep{secatscale,monna}. However, critical challenges remain unresolved. The primary obstacle is achieving centralized-comparable throughput in the model-parallel setting—while recent work has narrowed this gap, performance overhead remains significant. While individual properties have been validated at moderate scales, no system has yet demonstrated all required capabilities simultaneously at frontier model scale (100B+ parameters). These challenges, while substantial, do not appear to be fundamental theoretical barriers, suggesting Protocol Learning is technically feasible with focused research effort.
\end{framed}

\section{Protocol Learning} \label{sec:main}
As noted in Sec. \ref{sec:compute}, decentralized training has the ability to assemble several orders of magnitude larger compute capacity than any centralized actor and consequently the ability to train models orders of magnitude larger than any today. It is our view that in order to achieve this scale (i.e. for the system to obtain the underlying compute and data required to train state-of-the-art models), decentralized training must be directly incentivized. There are many approaches to incentivization, however we focus on systems where ownership is allocated proportional to the training contribution. This is appealing as it aligns incentives and results in a self-contained system. As there is real cost to contributing (the cost of compute), trainers will not contribute to models that are not high expected utility (or if they do, it will be with the understanding they will make no return). This in turn creates market forces that push contributors towards development of the highest utility models for the lowest cost, not only in algorithm and learning design but also in hardware setup and location (due to energy prices, operating costs, etc.). If a trainer can perform the same computation cheaper, their return is larger. This choice to incentivize via fractional ownership can be viewed as implementing the meta-objective of maximal utility for the system more broadly. 

\subsection{Protocol Models}
Partial ownership through verifiable credentials is essential for attracting sufficient computational power to Protocol Learning, however such a setup is quite novel. We do not specify technical details of such a system here (this will appear in subsequent work); but do note several technical properties as they directly impact the risk landscape. 

Consider a trained model where inference requires credentials allocated during training in proportion to computational contribution. These credentials must be transferable—allowing downstream users to access the model similarly to current API systems—but with a crucial difference: rather than residing on centralized infrastructure, the model is redundantly sharded across participating nodes. Participants follow a serving protocol to collectively facilitate inference requests.

Critically, for such credentials to have value (and hence be an incentive), the model must be unextractable from the protocol. This is essential as once training has been complete, highly optimized, centralized inference can be setup with a lower total cost basis than the distributed inference proposed. In this scenario there is little use for the original swarm, and hence no incentive to participate in training. We term models that implement such a mechanism, and are unextractable from the protocol, `Protocol Models'. The fundamental property of Protocol Models is that while they are collaboratively trainable, inference is only be possible within the protocol, and at no point can any actor, or group of actors, extract a usable version of the model without spending more compute power than would be required to train the model from scratch. 

The viability of this approach hinges on model unextractability: the impossibility of reconstructing the complete model outside the protocol. Without this guarantee, actors could extract the model post-training and establish centralized inference services with lower operational costs, undermining the economic incentives for participation in the original network. We term systems with this property "Protocol Models," defined by two fundamental characteristics:
\begin{enumerate}
    \item Multiple parties can trustlessly collaborate to train the model. 
    \item The full weight set can never be extracted from the protocol.
\end{enumerate}

These properties create a self-reinforcing system where the value of participation credentials directly depends on the protocol's integrity and continued operation. Meeting only property 1 is the 'volunteer' decentralized training case described in Sec. \ref{sec:feasublity}. Hence, Unextractibility (property 2) must be combined with a system that supports large-scale training, runs on an dynamic and heterogeneous swarm, and is highly communication efficient due to the low-bandwidth interconnects used. Finally as there is no assumption of good actors in Protocol Learning, decentralized compute verification must be implemented in order to ensure the operations performed are correct. Protocol Learning implementations will hence likely draw from a convergence of multiple sub-fields including standard large-scale Distributed Training, Federated Learning (FL), and the previously unrelated field of cryptography.    

\subsection{Compute Verification}
A core problem in the Protocol Learning setting is the need to verify gradients and activations; as computational contributions are rewarded, there is a clear incentive to attempt to falsify model operations and collect the reward without incurring any cost. In the ideal case, the worker is able to provide a proof of computation alongside the gradient itself which is able to be both cheaply produced and checked relative to the cost of the gradient calculation. Compactness is essential; as discussed in Sec. \ref{overhead} communication efficiency is a critical part of any Protocol Learning system. Hence in addition to being cheap computationally, the proof itself must significantly smaller (in bits) than the gradient itself in order to not introduce overhead, as such proofs must be transferred alongside the gradients themselves.

Currently there are no effective methods to produce such proofs for frontier model workloads. While there is promising early work \cite{pol}, the current development stack is not numerically consistent at the gradient level \cite{numericalinconsistency}. Non-determinism introduced by rounding errors and aggregation order results in an identical layer operating on identical data to produce slightly different outputs, making such proofs of learning difficult \cite{polproblems}. 

An alternative direction to proofs, which guarantee correctness, is to provide game-theoretic guarantees by requiring compute contributors to stake small amounts of capital prior to contributing a gradient step. Gradients are then only periodically checked (via simple recalculation, and accepting some tolerance), however if the gradient is found to be incorrect, the staked capital is destroyed. Such setups can be designed to make contributing bad `work' economically irrational, as well as incentivizing validators to check work via 'jackpots' \cite{pos,livepeer}, however come with the disadvantage that a contributor must not only provide compute, but also lock up capital in order to participate.

\section{The Decentralized Frontier Risk Landscape} \label{sec:frontierrisk}
As noted in Sec. \ref{sec:compute}, Protocol Learning has the ability to assemble several orders of magnitude larger compute capacity than any centralized actor. It is our view that there are a range of approaches able to combine incentivization with the rapidly advancing progress in decentralized training, to facilitate training runs larger than current frontier models. This would result in a new class of very large models, which would alter the landscape of frontier risk both due to scale and hence capability, and also novel governance and access. 

Standard (Centralized) Frontier Risk assumes a single actor trains, owns and distributes the model. Risks can be bucketed into two main themes separated along an axis of base model capability; misalignment \citep{alignmentsurvey}, and misuse \citep{misuseopenai}. Common misuse risk categories include; cybersecurity \citep{aicyberrisk}, persuasion \citep{influencerisk} and chemical and biological threats \citep{biorisk}, and are feasible with models today or are predicted to be feasible in the short term future. Misalignment risks largely revolve around more capable models, that have achieved some level of autonomy \citep{alignmentorigional}. While there is a large emerging body of work on frontier risk to date no analysis of Decentralized Frontier Risk (DFR) has been carried out. DFR specifically focuses on frontier models trained in a decentralized manner with decentralized governance, as in the Protocol Learning setting. Implicit in this definition is that incentives are present; we argue in Sec. \ref{sec:compute} incentives are required to reach the scale possible to train a frontier model in the decentralized setting. Our discussion here is limited to those portions of the AI risk landscape that are meaningfully altered by decentralized models. Nevertheless, many challenges raised in the broader AI safety literature remain relevant to Protocol Models.

\subsection{Organizational-Level Power Consolidation}
Frontier model value is becoming increasingly apparent. Such models are likely to form a base dependency for a large potion of software due to their improved generality and capability over all previous systems. Consider, for example, the `AI tutor' scenario, where frontier models are adopted within most levels of education to assist students learning. This is already occurring \citep{aitutor} and appears to be a major short term application. While clearly beneficial over a textbook for queries such as `can you explain integration by parts?' due to ability to be interactive, contextual and personalized, such systems would directly mould the worldview of students when answering questions such as `how does an oligopoly form? Are oligopolies a bad thing?'. Differently to traditional learning where worldviews are shaped by various competing sources, opinions and individuals, if AI tutors become ubiquitous, and are powered by a small number of base models, worldviews would largely be shaped by a single source. Furthermore, in the current scenario, the design of this source would not be public (see Sec. \ref{subsec:poison}). In contrast, in Protocol Learning the training setup and datamix are all known. 

The AI tutor is simply one example; in addition to effectively allowing the controlling entities to decide what is `true' for what will be a ubiquitous and widely used reference source, frontier models concentrate power at the company level simply by being hugely economically valuable and difficult to create \citep{concentrationrisk}. Such concentration of power at the organizational level is almost universally considered undesirable. 

The incentives driving centralized actors to enhance frontier model capabilities are both powerful and evident today. As described in \citep{catesrophicriskoverview}, this “AI race” dynamic encourages organizations to make short-term decisions guided by the fear of rivals advancing more quickly, often to the detriment of safe and responsible development. By contrast, Protocol Learning removes direct competitive pressures: the largest models require collaborative training, shared ownership, and fractional governance. Because the combined compute of multiple contributors can surpass that of any single dominant entity, competitive advantages recede once it becomes clear that cooperation yields superior results. Although those with greater compute still exert more influence, the concentration of power is diminished, ensuring that smaller participants retain meaningful access and involvement.

\subsection{Geopolitical Competition and Power Consolidation}\label{subsec:georisk}

Polarization of capabilities will also emerge at the state level. As models grow more advanced and training costs rise, smaller developed nations will find it increasingly difficult to train their own frontier models (see Sec. \ref{sec:compute}). Consequently, they will be relegated to consumers of models developed elsewhere, wielding only limited, indirect influence over their design and behavior. This presents a challenge, as the widespread integration of frontier models is likely to have a direct cultural impact. In the centralized case the only path to model ownership for such countries is for training costs to decrease and scaling to no longer result in meaningful model improvements. In contrast, decentralized training potentially allows such governments, and the citizens of such governments to contribute their compute to various base models and participate directly in partial ownership and design decisions, without the need for massive capital outlay and infrastructure buildup required for standalone model development. 

Military applications further amplify these concerns. Notably, OpenAI removed its previous commitment to refrain from using frontier models for military purposes \citep{militaryban}, and other organizations never had similar restrictions. The allure of a state-level frontier model potentially driving geopolitical friction as nations vie for control over both the models themselves and the underlying supply chains is clear. However, guaranteed access to strong base models mitigates some of these competitive pressures especially in the case where training setups and data mixes are known.

A final note is to consider the impact of regulation when ability to train competitive foundation models is highly skewed at the state level but use is not. Significant efforts are underway to regulate the capabilities made available in both centralized \cite{ftc} and opensource \cite{opensourcereg} FM development. As FM development is currently centered solely within the largest countries, the governments and regulatory bodies of those countries supersede the control of the organizations creating the models, and hence can implement such regulation. This is perhaps reasonable \cite{trustingov} for the country in which the model was developed, however for model users in other countries (which will be the majority of users), such governance was not chosen. 

\subsection{Transparency and Poisoned Models} \label{subsec:poison}
In the current scenario, training recipes, data mixes, architectures and other design decisions are closely guarded trade secrets and are not released to model users. Users are able to evaluate the immediate obvious utility of the model but have no grantees around fine-grained behaviour in specific scenarios. It is likely then for use to accrue to the most broadly \textit{accurate} and \textit{useful} model, even if this model contains unacceptable or dangerous behaviour in specific areas injected either maliciously via model poisoning techniques \citep{sleeperagents}, or are emergent and undetected. In the decentralized scenario, model design is public, the data mix is known, algorithms are known and any alterations are known, allowing informed analysis and use and removing this entire risk category. 

\subsection{Proliferation and Model Forking}
A vulnerability in fully open-source or leaked frontier models is the potential for malicious fine-tuning. When a model’s weights are fully available, safeguards can be stripped away and harmful behavior introduced at scale. Such `jailbroken' versions can proliferate widely, and the probability of misuse is significantly increased.

In Protocol Learning, the model creators collectively establish and maintain decentralized governance over the base model’s evolution. While still encouraging broad participation and permissionless innovation—key dynamics of the open-source ecosystem—Protocol Learning frameworks prevent wholesale extraction of model weights. This is an essential property to allow monetization and hence incentivization of model creation (see Property 2 in Sec. \ref{sec:main}). AS a consequence however, changes must occur within the established, decentralized governance protocols. This ensures that modifications are subject to community oversight and must adhere to agreed-upon standards of safety, alignment, and responsible use, effectively reducing the risk of widespread malicious fine-tuning without stifling the collaborative spirit that drives advancement in open-source communities. Rather than a single actor proceeding unilaterally, many participants must agree to facilitate such use-cases. 

\subsection{The “No-Off” Problem} \label{sec:nooff}
We view this as the core risk introduced by Protocol Learning. The existential threat/rogue AI risk \cite{catesrophicriskoverview} scenario may seem speculative however it has received serious recent concern and study \cite{bengiorouge,openaiagentic}. In a centralized scenario, servers can be unpowered and datacenters can be isolated at the direction of a small number of individuals. In a decentralized scenario, as long as some portion of the swarm sufficient to support the model size remains online, the underlying model continues to operate. In a scenario where the model has achieved a level of performance able to influence human actors—an ability not contingent on self-awareness or agency—and either alignment techniques have failed, trainers have specifically designed it to do so, or for some other reason, the model can continue to recruit compute into the swarm. Consequently it is significantly more difficult to stop an unaligned AGI in this scenario; all participants must agree to pull compute power from the network within a short time-frame. 

This challenge is amplified by incentive structures in which returns are tied to model performance. A highly capable and valuable model can generate significant economic returns, making participants reluctant to discontinue their involvement. Not only must existing contributors forego expected profits by ceasing participation, but prospective newcomers must also decline potentially lucrative opportunities to join.

The severity of the no-off problem depends on the form of work verification employed by the protocol. If game-theoretic verification is implemented it is reduced; if a large run is deemed by external actors to be dangerous, it would be possible to spend large amounts to derail the training run by repeatedly joining the run and contributing bad gradients. Such “model derailment attacks” would be economically irrational under normal circumstances, but they represent a potential emergency measure to halt an unaligned and dangerous model. However, if the protocol achieves near-perfect work verification at low cost, these derailment strategies become ineffective. In that case, external intervention beyond the digital sphere—such as physically disrupting communication infrastructure—may be the only remaining option.

In light of these considerations, we view the no-off problem as the most significant long-term risk associated with Protocol Learning.

\section{Related Work}

\subsection{Volunteer Networks and Collaborative Training}
The Hivemind project \cite{hivemind} aims to train very large models on thousands of volunteer nodes connected via the internet. Hivemind integrates the works of \cite{moshpitsgd,secatscale,decenmoe} into a single project and is hence fault tolerant, fast, and decentralized. Hivemind is volunteer based, with no incentives to participate, and anyone may extract the trained weights after training is complete. A related project, Petals \cite{petals} focuses on collaborative inference only, but has demonstrated up to 6 token/s for a 70B parameter LLM is possible building on the Hivemind infrastructure. Petals also introduces the concept of communication efficient learning and sharing of adapters that can be attached to the base models being served in order to alter behavior, without altering the underlying model.

\subsection{Federated Learning} \label{sec:FL}
Federated Learning (FL) \cite{fl,flsurvey} aims to train a global model on edge devices using local, non-shared data. Edge devices complete some training and the result is sent back to a centralized server, with the local models periodically refreshed. This requires a centralized server, assumes good actors and homogeneous workers, and is largely motivated by data privacy. FL typically assumes a model can be stored on device, however there is recent work on federated large models \cite{federatedfm,fedfmsurvey}, as well as incentivization to contribute data in the FL setting \cite{flincentive1,vfchain}. Another approach is to federate only lightweight adapters of CFMs \cite{feddat,fedadaptation}, a task more suited to FL. Additionally, the first works combining blockchains and deep learning also appear in the FL literature, largely focused on facilitating FL in the presence of bad actors \cite{fedblock,blockchainfedsurvey, blockchainfedsurvey2, blockchainfedcomittee}. While there is overlap, the goal of Federated Learning (to train models with private data) is fundamentally different to Protocol Learning. 

\subsection{Decentralized Federated Learning}
Decentralized Federated Learning (DFL) \cite{dflsurvey} is an emerging variant of FL that aims to both address the single-point-of-failure risks in the classic FL approach and increase node-node bandwidth utilization and hence overall efficiency. In DFL there is no distinction between edge node and server roles and nodes must be self-organizing \cite{dflsurvey2}. Models are trained on local private datasets, but rather than sending updates to a server, updates are distributed amongst peers directly via gossip protocols \cite{gossipgrad, dflseggossip} or similar. Fixed network topologies, models that can fit on each node, and trusted participants are typically assumed \cite{dfltopology, dftcoms, ipls}. DFL shares many of the goals of Protocol Learning however assumes good actors and is hence not incentivized, and is not motivated by training the largest models possible and ensuring decentralized ownership.

\section{Conclusion}
Protocol Learning represents an alternative approach to frontier model development. Our analysis indicates that the technical foundations for this paradigm—including distributed training, byzantine fault tolerance, and heterogeneous compute coordination—are more mature than commonly recognized. While challenges remain in unifying these elements into a single system, they appear to be engineering rather than theoretical obstacles.

The incentive structures inherent to Protocol Learning enable the assembly of compute resources orders of magnitude larger than current centralized approaches. This capability, while powerful, introduces novel governance challenges and safety considerations. Our examination of Decentralized Frontier Risk demonstrates that Protocol Learning could reduce several key risk factors: it mitigates the consolidation of power among centralized actors, democratizes access to frontier capabilities, and increases transparency through public training procedures. However, it also introduces new concerns, most notably the "No-Off Problem"—the challenge of halting a deployed model when control is distributed across many incentivized participants.

Protocol Learning offers a path toward more democratized and transparent AI development while potentially enabling unprecedented scale. Realizing these benefits, while managing the associated risks, requires thoughtful system design and extensive research. We hope this work catalyzes increased attention to both the technical and governance challenges of decentralized AI development.

\clearpage
\bibliographystyle{plainnat}
\bibliography{main}

\begin{thebibliography}{93}
\providecommand{\natexlab}[1]{#1}
\providecommand{\url}[1]{\texttt{#1}}
\expandafter\ifx\csname urlstyle\endcsname\relax
  \providecommand{\doi}[1]{doi: #1}\else
  \providecommand{\doi}{doi: \begingroup \urlstyle{rm}\Url}\fi

\bibitem[Achiam et~al.(2023)Achiam, Adler, Agarwal, Ahmad, Akkaya, Aleman, Almeida, Altenschmidt, Altman, Anadkat, et~al.]{gpt4}
Josh Achiam, Steven Adler, Sandhini Agarwal, Lama Ahmad, Ilge Akkaya, Florencia~Leoni Aleman, Diogo Almeida, Janko Altenschmidt, Sam Altman, Shyamal Anadkat, et~al.
\newblock Gpt-4 technical report.
\newblock \emph{arXiv preprint arXiv:2303.08774}, 2023.

\bibitem[Alistarh et~al.(2017)Alistarh, Grubic, Li, Tomioka, and Vojnovic]{qsgd}
Dan Alistarh, Demjan Grubic, Jerry Li, Ryota Tomioka, and Milan Vojnovic.
\newblock Qsgd: Communication-efficient sgd via gradient quantization and encoding.
\newblock \emph{Advances in neural information processing systems}, 30, 2017.

\bibitem[Baruch et~al.(2019)Baruch, Baruch, and Goldberg]{byzrobproblems2}
Gilad Baruch, Moran Baruch, and Yoav Goldberg.
\newblock A little is enough: Circumventing defenses for distributed learning.
\newblock \emph{Advances in Neural Information Processing Systems}, 32, 2019.

\bibitem[Beltr{\'a}n et~al.(2023)Beltr{\'a}n, P{\'e}rez, S{\'a}nchez, Bernal, Bovet, P{\'e}rez, P{\'e}rez, and Celdr{\'a}n]{dflsurvey2}
Enrique Tom{\'a}s~Mart{\'\i}nez Beltr{\'a}n, Mario~Quiles P{\'e}rez, Pedro Miguel~S{\'a}nchez S{\'a}nchez, Sergio~L{\'o}pez Bernal, G{\'e}r{\^o}me Bovet, Manuel~Gil P{\'e}rez, Gregorio~Mart{\'\i}nez P{\'e}rez, and Alberto~Huertas Celdr{\'a}n.
\newblock Decentralized federated learning: Fundamentals, state of the art, frameworks, trends, and challenges.
\newblock \emph{IEEE Communications Surveys \& Tutorials}, 2023.

\bibitem[Bengio(2023)]{bengiorouge}
Yoshua Bengio.
\newblock How {Rogue} {AIs} may {Arise}, 2023.
\newblock URL \url{https://yoshuabengio.org/2023/05/22/how-rogue-ais-may-arise/}.

\bibitem[Blanchard et~al.(2017)Blanchard, El~Mhamdi, Guerraoui, and Stainer]{byztraining}
Peva Blanchard, El~Mahdi El~Mhamdi, Rachid Guerraoui, and Julien Stainer.
\newblock Machine learning with adversaries: Byzantine tolerant gradient descent.
\newblock \emph{Advances in neural information processing systems}, 30, 2017.

\bibitem[Blot et~al.(2016)Blot, Picard, Cord, and Thome]{gossipdl}
Michael Blot, David Picard, Matthieu Cord, and Nicolas Thome.
\newblock Gossip training for deep learning.
\newblock \emph{arXiv preprint arXiv:1611.09726}, 2016.

\bibitem[Bonfanti(2022)]{aicyberrisk}
Matteo~E Bonfanti.
\newblock Artificial intelligence and the offence-defence balance in cyber security.
\newblock \emph{Cyber Security: Socio-Technological Uncertainty and Political Fragmentation. London: Routledge}, pages 64--79, 2022.

\bibitem[Borzunov et~al.(2022)Borzunov, Baranchuk, Dettmers, Ryabinin, Belkada, Chumachenko, Samygin, and Raffel]{petals}
Alexander Borzunov, Dmitry Baranchuk, Tim Dettmers, Max Ryabinin, Younes Belkada, Artem Chumachenko, Pavel Samygin, and Colin Raffel.
\newblock Petals: Collaborative inference and fine-tuning of large models.
\newblock \emph{arXiv preprint arXiv:2209.01188}, 2022.

\bibitem[Boyd et~al.(2006)Boyd, Ghosh, Prabhakar, and Shah]{gossipproto}
Stephen Boyd, Arpita Ghosh, Balaji Prabhakar, and Devavrat Shah.
\newblock Randomized gossip algorithms.
\newblock \emph{IEEE transactions on information theory}, 52\penalty0 (6):\penalty0 2508--2530, 2006.

\bibitem[Brown et~al.(2020)Brown, Mann, Ryder, Subbiah, Kaplan, Dhariwal, Neelakantan, Shyam, Sastry, Askell, et~al.]{gpt3}
Tom Brown, Benjamin Mann, Nick Ryder, Melanie Subbiah, Jared~D Kaplan, Prafulla Dhariwal, Arvind Neelakantan, Pranav Shyam, Girish Sastry, Amanda Askell, et~al.
\newblock Language models are few-shot learners.
\newblock \emph{Advances in neural information processing systems}, 33:\penalty0 1877--1901, 2020.

\bibitem[Brundage et~al.(2022)Brundage, Mayer, Eloundou, Agarwal, Adler, Krueger, Leike, and Mishkin]{misuseopenai}
Miles Brundage, Katie Mayer, Tyna Eloundou, Sandhini Agarwal, Steven Adler, Gretchen Krueger, Jan Leike, and Pamela Mishkin.
\newblock Lessons learned on language model safety and misuse, 2022.

\bibitem[Bubeck et~al.(2023)Bubeck, Chandrasekaran, Eldan, Gehrke, Horvitz, Kamar, Lee, Lee, Li, Lundberg, et~al.]{sparks}
S{\'e}bastien Bubeck, Varun Chandrasekaran, Ronen Eldan, Johannes Gehrke, Eric Horvitz, Ece Kamar, Peter Lee, Yin~Tat Lee, Yuanzhi Li, Scott Lundberg, et~al.
\newblock Sparks of artificial general intelligence: Early experiments with gpt-4.
\newblock \emph{arXiv preprint arXiv:2303.12712}, 2023.

\bibitem[Chen et~al.(2023)Chen, Zhang, Krompass, Gu, and Tresp]{feddat}
Haokun Chen, Yao Zhang, Denis Krompass, Jindong Gu, and Volker Tresp.
\newblock Feddat: An approach for foundation model finetuning in multi-modal heterogeneous federated learning.
\newblock \emph{arXiv preprint arXiv:2308.12305}, 2023.

\bibitem[Daily et~al.(2018)Daily, Vishnu, Siegel, Warfel, and Amatya]{gossipgrad}
Jeff Daily, Abhinav Vishnu, Charles Siegel, Thomas Warfel, and Vinay Amatya.
\newblock Gossipgrad: Scalable deep learning using gossip communication based asynchronous gradient descent.
\newblock \emph{arXiv preprint arXiv:1803.05880}, 2018.

\bibitem[Damaskinos et~al.(2019)Damaskinos, El-Mhamdi, Guerraoui, Guirguis, and Rouault]{Aggregathor}
Georgios Damaskinos, El-Mahdi El-Mhamdi, Rachid Guerraoui, Arsany Guirguis, and S{\'e}bastien Rouault.
\newblock Aggregathor: Byzantine machine learning via robust gradient aggregation.
\newblock \emph{Proceedings of Machine Learning and Systems}, 1:\penalty0 81--106, 2019.

\bibitem[Diskin et~al.(2021)Diskin, Bukhtiyarov, Ryabinin, Saulnier, Sinitsin, Popov, Pyrkin, Kashirin, Borzunov, Villanova~del Moral, et~al.]{opencollabs}
Michael Diskin, Alexey Bukhtiyarov, Max Ryabinin, Lucile Saulnier, Anton Sinitsin, Dmitry Popov, Dmitry~V Pyrkin, Maxim Kashirin, Alexander Borzunov, Albert Villanova~del Moral, et~al.
\newblock Distributed deep learning in open collaborations.
\newblock \emph{Advances in Neural Information Processing Systems}, 34:\penalty0 7879--7897, 2021.

\bibitem[El-Mhamdi et~al.(2020)El-Mhamdi, Guerraoui, Guirguis, Hoang, and Rouault]{el2020genuinely}
El-Mahdi El-Mhamdi, Rachid Guerraoui, Arsany Guirguis, L{\^e}~Nguy{\^e}n Hoang, and S{\'e}bastien Rouault.
\newblock Genuinely distributed byzantine machine learning.
\newblock In \emph{Proceedings of the 39th Symposium on Principles of Distributed Computing}, pages 355--364, 2020.

\bibitem[Faghri et~al.(2020)Faghri, Tabrizian, Markov, Alistarh, Roy, and Ramezani-Kebrya]{adaptivequantization}
Fartash Faghri, Iman Tabrizian, Ilia Markov, Dan Alistarh, Daniel~M Roy, and Ali Ramezani-Kebrya.
\newblock Adaptive gradient quantization for data-parallel sgd.
\newblock \emph{Advances in neural information processing systems}, 33:\penalty0 3174--3185, 2020.

\bibitem[Fang et~al.(2023)Fang, Jia, Thudi, Yaghini, Choquette-Choo, Dullerud, Chandrasekaran, and Papernot]{polproblems}
Congyu Fang, Hengrui Jia, Anvith Thudi, Mohammad Yaghini, Christopher~A Choquette-Choo, Natalie Dullerud, Varun Chandrasekaran, and Nicolas Papernot.
\newblock Proof-of-learning is currently more broken than you think.
\newblock In \emph{2023 IEEE 8th European Symposium on Security and Privacy (EuroS\&P)}, pages 797--816. IEEE, 2023.

\bibitem[Farhadkhani et~al.(2022)Farhadkhani, Guerraoui, Gupta, Hoang, Pinot, and Stephan]{monna}
Sadegh Farhadkhani, Rachid Guerraoui, Nirupam Gupta, L{\^e}~Nguy{\^e}n Hoang, Rafael Pinot, and John Stephan.
\newblock Making byzantine decentralized learning efficient.
\newblock \emph{arXiv preprint arXiv:2209.10931}, 2022.

\bibitem[Field(2024)]{militaryban}
Hayden Field.
\newblock {OpenAI} quietly removes ban on military use of its {AI} tools, January 2024.
\newblock URL \url{https://archive.md/cC9yp}.

\bibitem[FTC(2024)]{ftc}
FTC.
\newblock {FTC} {Launches} {Inquiry} into {Generative} {AI} {Investments} and {Partnerships}, January 2024.
\newblock URL \url{https://archive.md/3EzdQ}.

\bibitem[{Future of Life Institute}(2023)]{pauseletter}
{Future of Life Institute}.
\newblock Pause {Giant} {AI} {Experiments}: {An} {Open} {Letter}, 2023.
\newblock URL \url{https://futureoflife.org/open-letter/pause-giant-ai-experiments/}.

\bibitem[Gabrielli et~al.(2023)Gabrielli, Pica, and Tolomei]{dflsurvey}
Edoardo Gabrielli, Giovanni Pica, and Gabriele Tolomei.
\newblock A survey on decentralized federated learning.
\newblock \emph{arXiv preprint arXiv:2308.04604}, 2023.

\bibitem[Goldstein et~al.(2023)Goldstein, Sastry, Musser, DiResta, Gentzel, and Sedova]{influencerisk}
Josh~A Goldstein, Girish Sastry, Micah Musser, Renee DiResta, Matthew Gentzel, and Katerina Sedova.
\newblock Generative language models and automated influence operations: Emerging threats and potential mitigations.
\newblock \emph{arXiv preprint arXiv:2301.04246}, 2023.

\bibitem[Gorbunov et~al.(2022)Gorbunov, Borzunov, Diskin, and Ryabinin]{secatscale}
Eduard Gorbunov, Alexander Borzunov, Michael Diskin, and Max Ryabinin.
\newblock Secure distributed training at scale.
\newblock In \emph{International Conference on Machine Learning}, pages 7679--7739. PMLR, 2022.

\bibitem[Harlap et~al.(2018)Harlap, Narayanan, Phanishayee, Seshadri, Devanur, Ganger, and Gibbons]{pipeparrellel}
Aaron Harlap, Deepak Narayanan, Amar Phanishayee, Vivek Seshadri, Nikhil Devanur, Greg Ganger, and Phil Gibbons.
\newblock Pipedream: Fast and efficient pipeline parallel dnn training.
\newblock \emph{arXiv preprint arXiv:1806.03377}, 2018.

\bibitem[Harris(2024)]{opensourcereg}
David~Evan Harris.
\newblock Open-{Source} {AI} {Is} {Uniquely} {Dangerous} - {IEEE} {Spectrum}, 2024.
\newblock URL \url{https://spectrum.ieee.org/open-source-ai-2666932122}.

\bibitem[Hendrycks et~al.(2023)Hendrycks, Mazeika, and Woodside]{catesrophicriskoverview}
Dan Hendrycks, Mantas Mazeika, and Thomas Woodside.
\newblock An overview of catastrophic ai risks.
\newblock \emph{arXiv preprint arXiv:2306.12001}, 2023.

\bibitem[Hoffmann et~al.(2022)Hoffmann, Borgeaud, Mensch, Buchatskaya, Cai, Rutherford, Casas, Hendricks, Welbl, Clark, et~al.]{scalinglaws}
Jordan Hoffmann, Sebastian Borgeaud, Arthur Mensch, Elena Buchatskaya, Trevor Cai, Eliza Rutherford, Diego de~Las Casas, Lisa~Anne Hendricks, Johannes Welbl, Aidan Clark, et~al.
\newblock Training compute-optimal large language models.
\newblock \emph{arXiv preprint arXiv:2203.15556}, 2022.

\bibitem[Hu et~al.(2019)Hu, Jiang, and Wang]{dflseggossip}
Chenghao Hu, Jingyan Jiang, and Zhi Wang.
\newblock Decentralized federated learning: A segmented gossip approach.
\newblock \emph{arXiv preprint arXiv:1908.07782}, 2019.

\bibitem[Huang et~al.(2019)Huang, Cheng, Bapna, Firat, Chen, Chen, Lee, Ngiam, Le, Wu, et~al.]{gpipe}
Yanping Huang, Youlong Cheng, Ankur Bapna, Orhan Firat, Dehao Chen, Mia Chen, HyoukJoong Lee, Jiquan Ngiam, Quoc~V Le, Yonghui Wu, et~al.
\newblock Gpipe: Efficient training of giant neural networks using pipeline parallelism.
\newblock \emph{Advances in neural information processing systems}, 32, 2019.

\bibitem[Hubinger et~al.(2024)Hubinger, Denison, Mu, Lambert, Tong, MacDiarmid, Lanham, Ziegler, Maxwell, Cheng, et~al.]{sleeperagents}
Evan Hubinger, Carson Denison, Jesse Mu, Mike Lambert, Meg Tong, Monte MacDiarmid, Tamera Lanham, Daniel~M Ziegler, Tim Maxwell, Newton Cheng, et~al.
\newblock Sleeper agents: Training deceptive llms that persist through safety training.
\newblock \emph{arXiv preprint arXiv:2401.05566}, 2024.

\bibitem[Ji et~al.(2023)Ji, Qiu, Chen, Zhang, Lou, Wang, Duan, He, Zhou, Zhang, et~al.]{alignmentsurvey}
Jiaming Ji, Tianyi Qiu, Boyuan Chen, Borong Zhang, Hantao Lou, Kaile Wang, Yawen Duan, Zhonghao He, Jiayi Zhou, Zhaowei Zhang, et~al.
\newblock Ai alignment: A comprehensive survey.
\newblock \emph{arXiv preprint arXiv:2310.19852}, 2023.

\bibitem[Jia et~al.(2021)Jia, Yaghini, Choquette-Choo, Dullerud, Thudi, Chandrasekaran, and Papernot]{pol}
Hengrui Jia, Mohammad Yaghini, Christopher~A Choquette-Choo, Natalie Dullerud, Anvith Thudi, Varun Chandrasekaran, and Nicolas Papernot.
\newblock Proof-of-learning: Definitions and practice.
\newblock In \emph{2021 IEEE Symposium on Security and Privacy (SP)}, pages 1039--1056. IEEE, 2021.

\bibitem[Kairouz et~al.(2021)Kairouz, McMahan, Avent, Bellet, Bennis, Bhagoji, Bonawitz, Charles, Cormode, Cummings, et~al.]{flsurvey}
Peter Kairouz, H~Brendan McMahan, Brendan Avent, Aur{\'e}lien Bellet, Mehdi Bennis, Arjun~Nitin Bhagoji, Kallista Bonawitz, Zachary Charles, Graham Cormode, Rachel Cummings, et~al.
\newblock Advances and open problems in federated learning.
\newblock \emph{Foundations and Trends{\textregistered} in Machine Learning}, 14\penalty0 (1--2):\penalty0 1--210, 2021.

\bibitem[Kang et~al.(2019)Kang, Xiong, Niyato, Yu, Liang, and Kim]{flincentive1}
Jiawen Kang, Zehui Xiong, Dusit Niyato, Han Yu, Ying-Chang Liang, and Dong~In Kim.
\newblock Incentive design for efficient federated learning in mobile networks: A contract theory approach.
\newblock In \emph{2019 IEEE VTS Asia Pacific Wireless Communications Symposium (APWCS)}, pages 1--5. IEEE, 2019.

\bibitem[Kang et~al.(2023)Kang, Fan, Gu, Fan, and Yang]{fedadaptation}
Yan Kang, Tao Fan, Hanlin Gu, Lixin Fan, and Qiang Yang.
\newblock Grounding foundation models through federated transfer learning: A general framework.
\newblock \emph{arXiv preprint arXiv:2311.17431}, 2023.

\bibitem[Karimireddy et~al.(2021)Karimireddy, He, and Jaggi]{centeredclip}
Sai~Praneeth Karimireddy, Lie He, and Martin Jaggi.
\newblock Learning from history for byzantine robust optimization.
\newblock In \emph{International Conference on Machine Learning}, pages 5311--5319. PMLR, 2021.

\bibitem[King and Nadal(2012)]{pos}
Sunny King and Scott Nadal.
\newblock Ppcoin: Peer-to-peer crypto-currency with proof-of-stake.
\newblock \emph{self-published paper, August}, 19\penalty0 (1), 2012.

\bibitem[Koloskova et~al.(2020)Koloskova, Loizou, Boreiri, Jaggi, and Stich]{unifieddsgd}
Anastasia Koloskova, Nicolas Loizou, Sadra Boreiri, Martin Jaggi, and Sebastian Stich.
\newblock A unified theory of decentralized sgd with changing topology and local updates.
\newblock In \emph{International Conference on Machine Learning}, pages 5381--5393. PMLR, 2020.

\bibitem[Krizhevsky et~al.(2012)Krizhevsky, Sutskever, and Hinton]{dp}
Alex Krizhevsky, Ilya Sutskever, and Geoffrey~E Hinton.
\newblock Imagenet classification with deep convolutional neural networks.
\newblock \emph{Advances in neural information processing systems}, 25, 2012.

\bibitem[Larson et~al.(2009)Larson, Snow, Shirts, and Pande]{foldingathome}
Stefan~M Larson, Christopher~D Snow, Michael Shirts, and Vijay~S Pande.
\newblock Folding at home and genome at home: Using distributed computing to tackle previously intractable problems in computational biology.
\newblock \emph{arXiv preprint arXiv:0901.0866}, 2009.

\bibitem[Learning{@}home-Team(2020)]{hivemind}
Learning{@}home-Team.
\newblock {H}ivemind: a {L}ibrary for {D}ecentralized {D}eep {L}earning.
\newblock \url{https://github.com/learning-at-home/hivemind}, 2020.

\bibitem[Li et~al.(2014)Li, Andersen, Park, Smola, Ahmed, Josifovski, Long, Shekita, and Su]{ps}
Mu~Li, David~G Andersen, Jun~Woo Park, Alexander~J Smola, Amr Ahmed, Vanja Josifovski, James Long, Eugene~J Shekita, and Bor-Yiing Su.
\newblock Scaling distributed machine learning with the parameter server.
\newblock In \emph{11th USENIX Symposium on operating systems design and implementation (OSDI 14)}, pages 583--598, 2014.

\bibitem[Li et~al.(2020{\natexlab{a}})Li, Zhao, Varma, Salpekar, Noordhuis, Li, Paszke, Smith, Vaughan, Damania, et~al.]{ddppytorch}
Shen Li, Yanli Zhao, Rohan Varma, Omkar Salpekar, Pieter Noordhuis, Teng Li, Adam Paszke, Jeff Smith, Brian Vaughan, Pritam Damania, et~al.
\newblock Pytorch distributed: Experiences on accelerating data parallel training.
\newblock \emph{arXiv preprint arXiv:2006.15704}, 2020{\natexlab{a}}.

\bibitem[Li et~al.(2020{\natexlab{b}})Li, Chen, Liu, Huang, Zheng, and Yan]{blockchainfedcomittee}
Yuzheng Li, Chuan Chen, Nan Liu, Huawei Huang, Zibin Zheng, and Qiang Yan.
\newblock A blockchain-based decentralized federated learning framework with committee consensus.
\newblock \emph{IEEE Network}, 35\penalty0 (1):\penalty0 234--241, 2020{\natexlab{b}}.

\bibitem[Li et~al.(2020{\natexlab{c}})Li, Chen, Liu, Huang, Zheng, and Yan]{fedblock}
Yuzheng Li, Chuan Chen, Nan Liu, Huawei Huang, Zibin Zheng, and Qiang Yan.
\newblock A blockchain-based decentralized federated learning framework with committee consensus.
\newblock \emph{IEEE Network}, 35\penalty0 (1):\penalty0 234--241, 2020{\natexlab{c}}.

\bibitem[Li et~al.(2017)Li, Davis, and Jarvis]{butterflyallreduce}
Zhenyu Li, James Davis, and Stephen Jarvis.
\newblock An efficient task-based all-reduce for machine learning applications.
\newblock In \emph{Proceedings of the Machine Learning on HPC Environments}, pages 1--8. 2017.

\bibitem[Lian et~al.(2017)Lian, Zhang, Zhang, Hsieh, Zhang, and Liu]{D-PSGD}
Xiangru Lian, Ce~Zhang, Huan Zhang, Cho-Jui Hsieh, Wei Zhang, and Ji~Liu.
\newblock Can decentralized algorithms outperform centralized algorithms? a case study for decentralized parallel stochastic gradient descent.
\newblock \emph{Advances in neural information processing systems}, 30, 2017.

\bibitem[Lian et~al.(2018)Lian, Zhang, Zhang, and Liu]{AD-PSGD}
Xiangru Lian, Wei Zhang, Ce~Zhang, and Ji~Liu.
\newblock Asynchronous decentralized parallel stochastic gradient descent.
\newblock In \emph{International Conference on Machine Learning}, pages 3043--3052. PMLR, 2018.

\bibitem[Liu et~al.(2022)Liu, Chen, and Zhang]{dftcoms}
Wei Liu, Li~Chen, and Wenyi Zhang.
\newblock Decentralized federated learning: Balancing communication and computing costs.
\newblock \emph{IEEE Transactions on Signal and Information Processing over Networks}, 8:\penalty0 131--143, 2022.

\bibitem[Markov et~al.(2023)Markov, Vladu, Guo, and Alistarh]{qfsdp}
Ilia Markov, Adrian Vladu, Qi~Guo, and Dan Alistarh.
\newblock Quantized distributed training of large models with convergence guarantees.
\newblock ICML'23. JMLR.org, 2023.

\bibitem[McMahan et~al.(2017)McMahan, Moore, Ramage, Hampson, and y~Arcas]{fl}
Brendan McMahan, Eider Moore, Daniel Ramage, Seth Hampson, and Blaise~Aguera y~Arcas.
\newblock Communication-efficient learning of deep networks from decentralized data.
\newblock In \emph{Artificial intelligence and statistics}, pages 1273--1282. PMLR, 2017.

\bibitem[Messina(2023)]{bitcoinenergy}
Irene Messina.
\newblock Bitcoin electricity consumption: an improved assessment - {News} \& insight, August 2023.
\newblock URL \url{https://www.jbs.cam.ac.uk/2023/bitcoin-electricity-consumption/}.

\bibitem[Mhamdi et~al.(2018)Mhamdi, Guerraoui, and Rouault]{byzrobproblems}
El~Mahdi~El Mhamdi, Rachid Guerraoui, and S{\'e}bastien Rouault.
\newblock The hidden vulnerability of distributed learning in byzantium.
\newblock \emph{arXiv preprint arXiv:1802.07927}, 2018.

\bibitem[Nakamoto(2008)]{bitcoin}
Satoshi Nakamoto.
\newblock Bitcoin: A peer-to-peer electronic cash system.
\newblock \emph{Decentralized business review}, 2008.

\bibitem[Nguyen et~al.(2022)Nguyen, Frey, Gonz{\'a}lez, and Brezzi]{trustingov}
David Nguyen, Val{\'e}rie Frey, Santiago Gonz{\'a}lez, and Monica Brezzi.
\newblock Survey design and technical documentation supporting the 2021 oecd survey on drivers of trust in government institutions.
\newblock 2022.

\bibitem[NVIDIA(2023)]{h100specs}
NVIDIA.
\newblock {NVIDIA} {H100} {Tensor} {Core} {GPU} {Architecture} {Overview}, 2023.
\newblock URL \url{https://resources.nvidia.com/en-us-tensor-core}.

\bibitem[Owen(2024)]{predictablescaling}
David Owen.
\newblock How predictable is language model benchmark performance?, 2024.

\bibitem[Pappas et~al.(2021)Pappas, Chatzopoulos, Lalis, and Vavalis]{ipls}
Christodoulos Pappas, Dimitris Chatzopoulos, Spyros Lalis, and Manolis Vavalis.
\newblock Ipls: A framework for decentralized federated learning.
\newblock In \emph{2021 IFIP Networking Conference (IFIP Networking)}, pages 1--6. IEEE, 2021.

\bibitem[Patarasuk and Yuan(2009)]{allreducevariants}
Pitch Patarasuk and Xin Yuan.
\newblock Bandwidth optimal all-reduce algorithms for clusters of workstations.
\newblock \emph{Journal of Parallel and Distributed Computing}, 69\penalty0 (2):\penalty0 117--124, 2009.

\bibitem[Peng et~al.(2021{\natexlab{a}})Peng, Li, and Ling]{peng2021}
Jie Peng, Weiyu Li, and Qing Ling.
\newblock Byzantine-robust decentralized stochastic optimization over static and time-varying networks.
\newblock \emph{Signal Processing}, 183:\penalty0 108020, 2021{\natexlab{a}}.

\bibitem[Peng et~al.(2021{\natexlab{b}})Peng, Xu, Chu, Gao, Yao, Gu, and Tang]{vfchain}
Zhe Peng, Jianliang Xu, Xiaowen Chu, Shang Gao, Yuan Yao, Rong Gu, and Yuzhe Tang.
\newblock Vfchain: Enabling verifiable and auditable federated learning via blockchain systems.
\newblock \emph{IEEE Transactions on Network Science and Engineering}, 9\penalty0 (1):\penalty0 173--186, 2021{\natexlab{b}}.

\bibitem[Petkanics and Tang(2018)]{livepeer}
Doug Petkanics and Eric Tang.
\newblock Livepeer whitepaper.
\newblock \emph{Technical report, Livepeer}, 2018.

\bibitem[Qammar et~al.(2023)Qammar, Karim, Ning, and Ding]{blockchainfedsurvey2}
Attia Qammar, Ahmad Karim, Huansheng Ning, and Jianguo Ding.
\newblock Securing federated learning with blockchain: a systematic literature review.
\newblock \emph{Artificial Intelligence Review}, 56\penalty0 (5):\penalty0 3951--3985, 2023.

\bibitem[Ren et~al.(2021)Ren, Rajbhandari, Aminabadi, Ruwase, Yang, Zhang, Li, and He]{zero3}
Jie Ren, Samyam Rajbhandari, Reza~Yazdani Aminabadi, Olatunji Ruwase, Shuangyan Yang, Minjia Zhang, Dong Li, and Yuxiong He.
\newblock $\{$ZeRO-Offload$\}$: Democratizing $\{$Billion-Scale$\}$ model training.
\newblock In \emph{2021 USENIX Annual Technical Conference (USENIX ATC 21)}, pages 551--564, 2021.

\bibitem[Ryabinin and Gusev(2020)]{decenmoe}
Max Ryabinin and Anton Gusev.
\newblock Towards crowdsourced training of large neural networks using decentralized mixture-of-experts.
\newblock \emph{Advances in Neural Information Processing Systems}, 33:\penalty0 3659--3672, 2020.

\bibitem[Ryabinin et~al.(2021)Ryabinin, Gorbunov, Plokhotnyuk, and Pekhimenko]{moshpitsgd}
Max Ryabinin, Eduard Gorbunov, Vsevolod Plokhotnyuk, and Gennady Pekhimenko.
\newblock Moshpit sgd: Communication-efficient decentralized training on heterogeneous unreliable devices.
\newblock \emph{Advances in Neural Information Processing Systems}, 34:\penalty0 18195--18211, 2021.

\bibitem[Ryabinin et~al.(2023)Ryabinin, Dettmers, Diskin, and Borzunov]{swarmparellel}
Max Ryabinin, Tim Dettmers, Michael Diskin, and Alexander Borzunov.
\newblock Swarm parallelism: Training large models can be surprisingly communication-efficient.
\newblock \emph{arXiv preprint arXiv:2301.11913}, 2023.

\bibitem[Schaeffer et~al.(2023)Schaeffer, Miranda, and Koyejo]{emergentmirage}
Rylan Schaeffer, Brando Miranda, and Sanmi Koyejo.
\newblock Are emergent abilities of large language models a mirage?
\newblock \emph{arXiv preprint arXiv:2304.15004}, 2023.

\bibitem[Schl{\"o}gl et~al.(2024)Schl{\"o}gl, Hofer, and B{\"o}hme]{numericalinconsistency}
Alex Schl{\"o}gl, Nora Hofer, and Rainer B{\"o}hme.
\newblock Causes and effects of unanticipated numerical deviations in neural network inference frameworks.
\newblock \emph{Advances in Neural Information Processing Systems}, 36, 2024.

\bibitem[Sergeev and Del~Balso(2018)]{hovrod}
Alexander Sergeev and Mike Del~Balso.
\newblock Horovod: fast and easy distributed deep learning in tensorflow.
\newblock \emph{arXiv preprint arXiv:1802.05799}, 2018.

\bibitem[Shavit et~al.(2023)Shavit, Agarwal, Brundage, Adler, O’Keefe, Campbell, Lee, Mishkin, Eloundou, Hickey, et~al.]{openaiagentic}
Yonadav Shavit, Sandhini Agarwal, Miles Brundage, Steven Adler, Cullen O’Keefe, Rosie Campbell, Teddy Lee, Pamela Mishkin, Tyna Eloundou, Alan Hickey, et~al.
\newblock Practices for governing agentic ai systems.
\newblock 2023.

\bibitem[Shoeybi et~al.(2019)Shoeybi, Patwary, Puri, LeGresley, Casper, and Catanzaro]{megatronmodelparallel}
Mohammad Shoeybi, Mostofa Patwary, Raul Puri, Patrick LeGresley, Jared Casper, and Bryan Catanzaro.
\newblock Megatron-lm: Training multi-billion parameter language models using model parallelism.
\newblock \emph{arXiv preprint arXiv:1909.08053}, 2019.

\bibitem[Tang et~al.(2020{\natexlab{a}})Tang, Shi, and Chu]{dladaptive}
Zhenheng Tang, Shaohuai Shi, and Xiaowen Chu.
\newblock Communication-efficient decentralized learning with sparsification and adaptive peer selection.
\newblock In \emph{2020 IEEE 40th International Conference on Distributed Computing Systems (ICDCS)}, pages 1207--1208. IEEE, 2020{\natexlab{a}}.

\bibitem[Tang et~al.(2020{\natexlab{b}})Tang, Shi, Wang, Li, and Chu]{qsgdsummary}
Zhenheng Tang, Shaohuai Shi, Wei Wang, Bo~Li, and Xiaowen Chu.
\newblock Communication-efficient distributed deep learning: A comprehensive survey.
\newblock \emph{arXiv preprint arXiv:2003.06307}, 2020{\natexlab{b}}.

\bibitem[Urbina et~al.(2022)Urbina, Lentzos, Invernizzi, and Ekins]{biorisk}
Fabio Urbina, Filippa Lentzos, C{\'e}dric Invernizzi, and Sean Ekins.
\newblock Dual use of artificial-intelligence-powered drug discovery.
\newblock \emph{Nature Machine Intelligence}, 4\penalty0 (3):\penalty0 189--191, 2022.

\bibitem[Vanian(2024)]{fbpurchase}
Jonathan Vanian.
\newblock Mark {Zuckerberg} indicates {Meta} is spending billions of dollars on {Nvidia} {AI} chips, January 2024.
\newblock URL \url{https://archive.md/UqV1x}.

\bibitem[Vaswani et~al.(2017)Vaswani, Shazeer, Parmar, Uszkoreit, Jones, Gomez, Kaiser, and Polosukhin]{transformer}
Ashish Vaswani, Noam Shazeer, Niki Parmar, Jakob Uszkoreit, Llion Jones, Aidan~N Gomez, {\L}ukasz Kaiser, and Illia Polosukhin.
\newblock Attention is all you need.
\newblock \emph{Advances in neural information processing systems}, 30, 2017.

\bibitem[Vipra and Korinek(2023)]{concentrationrisk}
Jai Vipra and Anton Korinek.
\newblock Market concentration implications of foundation models.
\newblock \emph{arXiv preprint arXiv:2311.01550}, 2023.

\bibitem[Wang et~al.(2021)Wang, Liu, Zheng, Dai, Jia, and Xie]{dfltopology}
Tian Wang, Yan Liu, Xi~Zheng, Hong-Ning Dai, Weijia Jia, and Mande Xie.
\newblock Edge-based communication optimization for distributed federated learning.
\newblock \emph{IEEE Transactions on Network Science and Engineering}, 9\penalty0 (4):\penalty0 2015--2024, 2021.

\bibitem[Wei et~al.(2022)Wei, Tay, Bommasani, Raffel, Zoph, Borgeaud, Yogatama, Bosma, Zhou, Metzler, et~al.]{emergentprops}
Jason Wei, Yi~Tay, Rishi Bommasani, Colin Raffel, Barret Zoph, Sebastian Borgeaud, Dani Yogatama, Maarten Bosma, Denny Zhou, Donald Metzler, et~al.
\newblock Emergent abilities of large language models.
\newblock \emph{arXiv preprint arXiv:2206.07682}, 2022.

\bibitem[Wiener(1960)]{alignmentorigional}
Norbert Wiener.
\newblock Some moral and technical consequences of automation: As machines learn they may develop unforeseen strategies at rates that baffle their programmers.
\newblock \emph{Science}, 131\penalty0 (3410):\penalty0 1355--1358, 1960.

\bibitem[Wood(2014)]{ethereum}
Gavin Wood.
\newblock Ethereum: A secure decentralised generalised transaction ledger.
\newblock 2014.

\bibitem[Xie et~al.(2020)Xie, Koyejo, and Gupta]{byzrobproblems3}
Cong Xie, Oluwasanmi Koyejo, and Indranil Gupta.
\newblock Fall of empires: Breaking byzantine-tolerant sgd by inner product manipulation.
\newblock In \emph{Uncertainty in Artificial Intelligence}, pages 261--270. PMLR, 2020.

\bibitem[Yesir and Rawat(2023)]{aitutor}
Ismail Yesir and Danda~B Rawat.
\newblock Recent advances in artificial intelligence enabled tutoring systems: A survey.
\newblock In \emph{2023 IEEE 13th Annual Computing and Communication Workshop and Conference (CCWC)}, pages 0375--0381. IEEE, 2023.

\bibitem[Yin et~al.(2018)Yin, Chen, Kannan, and Bartlett]{byzrob2}
Dong Yin, Yudong Chen, Ramchandran Kannan, and Peter Bartlett.
\newblock Byzantine-robust distributed learning: Towards optimal statistical rates.
\newblock In \emph{International Conference on Machine Learning}, pages 5650--5659. PMLR, 2018.

\bibitem[Yu et~al.(2023)Yu, Mu{\~n}oz, and Jannesari]{federatedfm}
Sixing Yu, J~Pablo Mu{\~n}oz, and Ali Jannesari.
\newblock Federated foundation models: Privacy-preserving and collaborative learning for large models.
\newblock \emph{arXiv preprint arXiv:2305.11414}, 2023.

\bibitem[Zhao et~al.(2023)Zhao, Gu, Varma, Luo, Huang, Xu, Wright, Shojanazeri, Ott, Shleifer, et~al.]{fsdp}
Yanli Zhao, Andrew Gu, Rohan Varma, Liang Luo, Chien-Chin Huang, Min Xu, Less Wright, Hamid Shojanazeri, Myle Ott, Sam Shleifer, et~al.
\newblock Pytorch fsdp: experiences on scaling fully sharded data parallel.
\newblock \emph{arXiv preprint arXiv:2304.11277}, 2023.

\bibitem[Zhu et~al.(2023)Zhu, Cao, Saxena, Jiang, and Ferradi]{blockchainfedsurvey}
Juncen Zhu, Jiannong Cao, Divya Saxena, Shan Jiang, and Houda Ferradi.
\newblock Blockchain-empowered federated learning: Challenges, solutions, and future directions.
\newblock \emph{ACM Computing Surveys}, 55\penalty0 (11):\penalty0 1--31, 2023.

\bibitem[Zhuang et~al.(2023)Zhuang, Chen, and Lyu]{fedfmsurvey}
Weiming Zhuang, Chen Chen, and Lingjuan Lyu.
\newblock When foundation model meets federated learning: Motivations, challenges, and future directions.
\newblock \emph{arXiv preprint arXiv:2306.15546}, 2023.

\end{thebibliography}

\newpage
\appendix
\onecolumn

\end{document}